\documentclass[10pt,twocolumn,letterpaper]{article}

\usepackage{cvpr}
\usepackage{times}
\usepackage{epsfig}
\usepackage{graphicx}
\usepackage{amsmath}
\usepackage{amssymb}

\usepackage{interval}

\usepackage{caption}

\usepackage{makecell}
\usepackage{booktabs}
\usepackage{multirow}
\usepackage{siunitx}

\usepackage[T1]{fontenc}
\usepackage{fix-cm}

\usepackage{times}
\usepackage{epsfig}
\usepackage{graphicx}
\usepackage{amsmath}
\usepackage{amssymb}

\usepackage{enumitem}

\usepackage{array}

\usepackage{caption}

\usepackage{makecell}
\usepackage{booktabs}
\usepackage{multirow}
\usepackage{siunitx}

\usepackage{fix-cm}

\usepackage{pifont}
\newcommand{\cmark}{\ding{51}}%
\newcommand{\xmark}{\ding{55}}%

\newcolumntype{?}{!{\vrule width 1pt}}
\newcolumntype{C}[1]{>{\centering}m{#1}}

\newcolumntype{X}{@{\hskip\tabcolsep\vrule width 1.5pt\hskip\tabcolsep}}

\newcommand{\myfigurethreecol}[1]{
\begin{minipage}[b]{.14\textwidth}
\includegraphics[width=1.1\linewidth]{#1}
\end{minipage}
}

\newcommand{\myfigurethreecolcaption}[2]{
\begin{minipage}[b]{.14\textwidth}
\includegraphics[width=1.1\linewidth]{#1}
\caption{{\scriptsize {#2}}}
\end{minipage}
}

\usepackage[pagebackref=true,breaklinks=true,letterpaper=true,colorlinks,bookmarks=false]{hyperref}

\cvprfinalcopy 

\def\cvprPaperID{282} 

\ifcvprfinal\fi
\begin{document}

\title{Convolutional Random Walk Networks for Semantic Image Segmentation}

\author{Gedas Bertasius$^{1}$, Lorenzo Torresani$^2$, Stella X. Yu$^3$, Jianbo Shi$^1$\\
$^1$University of Pennsylvania, $^2$Dartmouth College, $^3$UC Berkeley ICSI\\
{\tt\small \{gberta,jshi\}@seas.upenn.edu} \ \ \ {\tt\small lt@dartmouth.edu}  \ \ \ {\tt\small stella.yu@berkeley.edu}
}


\maketitle

\begin{abstract}


Most current semantic segmentation methods rely on fully convolutional networks (FCNs). However, their use of large receptive fields and many pooling layers cause low spatial resolution inside the deep layers. This leads to  predictions with poor localization around the boundaries. Prior work has attempted to address this issue by post-processing predictions with CRFs or MRFs. But such models often fail to capture semantic relationships between objects, which causes spatially disjoint predictions. To overcome these problems, recent methods integrated CRFs or MRFs into an FCN framework. The downside of these new models is that they have much higher complexity than traditional FCNs, which renders training and testing more challenging.

In this work we introduce a simple, yet effective Convolutional Random Walk Network (RWN) that addresses the issues of poor boundary localization and spatially fragmented predictions with very little increase in model complexity. Our proposed RWN jointly optimizes the objectives of pixelwise affinity and semantic segmentation. It combines these two objectives via a novel random walk layer that enforces consistent spatial grouping in the deep layers of the network. Our RWN is implemented using standard convolution and matrix multiplication. This allows an easy integration into existing FCN frameworks and it enables end-to-end training of the whole network via  standard back-propagation. Our implementation of RWN requires just $131$ additional parameters compared to the traditional FCNs, and yet it consistently produces an improvement over the FCNs on semantic segmentation and scene labeling. 




\end{abstract}

\vspace{-0.5cm}


\section{Introduction}

Fully convolutional networks (FCNs) were first introduced in~\cite{long_shelhamer_fcn} where they were shown to yield significant improvements in semantic image segmentation. Adopting the FCN approach, many subsequent methods have achieved even better performance ~\cite{DBLP:journals/corr/ChenPKMY14,crfasrnn_iccv2015,DBLP:journals/corr/DaiH015,hong2015decoupled,DBLP:journals/corr/LinSRH15,DBLP:journals/corr/ChenPKMY14,crfasrnn_iccv2015,DBLP:journals/corr/DaiH015,DBLP:journals/corr/LiuLLLT15,noh2015learning}. However, traditional FCN-based methods tend to suffer from several limitations. Large receptive fields in the convolutional layers and the presence of pooling layers lead to low spatial resolution in the deepest FCN layers. As a result, their predicted segments tend to be blobby and lack fine object boundary details. We report in Fig.~\ref{intro_fig} some examples illustrating typical poor localization of objects in the outputs of FCNs. Recently, Chen at al.~\cite{DBLP:journals/corr/ChenPKMY14} addressed this issue by applying a Dense-CRF post-processing step~\cite{NIPS2011_4296} on top of coarse FCN segmentations. However, such approaches often fail to accurately capture semantic relationships between objects and lead to spatially fragmented segmentations (see an example in the last column of Fig.~\ref{intro_fig}).

\captionsetup{labelformat=empty}
\captionsetup[figure]{skip=5pt}

\begin{figure}
\centering

\myfigurethreecolcaption{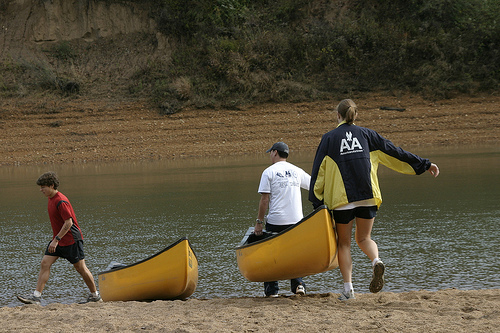}{Input}
\myfigurethreecolcaption{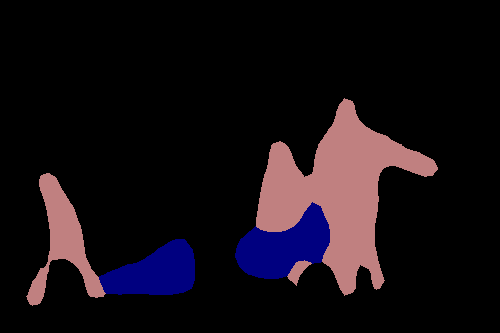}{DeepLab~\cite{DBLP:journals/corr/ChenPKMY14}}
\myfigurethreecolcaption{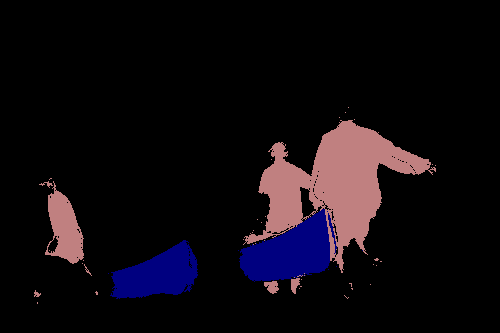}{DeepLab-CRF~\cite{DBLP:journals/corr/ChenPKMY14}}

\captionsetup{labelformat=default}
\setcounter{figure}{0}
   \caption{Examples illustrating shortcomings of prior semantic segmentation methods. Segments produced by FCNs are poorly localized around object boundaries, while Dense-CRF produce spatially-disjoint object segments. \vspace{-0.5cm}}
    \label{intro_fig}
\end{figure}

To address these problems several recent methods integrated CRFs or MRFs directly into the FCN framework~\cite{crfasrnn_iccv2015,DBLP:journals/corr/LiuLLLT15,DBLP:journals/corr/LinSRH15,Chen2016}. However, these new models typically involve (1) a large number of parameters, (2) complex loss functions requiring specialized model training or (3) recurrent layers, which make training and testing more complicated.  We summarize the most prominent of these approaches and their model complexities in Table~\ref{complexity_table}. 

We note that we do not claim that using complex loss functions always makes the model overly complex and too difficult to use. If a complex loss is integrated into an FCN framework such that the FCN can still be trained in a standard fashion, and produce better results than using standard losses, then such a model is beneficial. However, in the context of prior segmentation methods~\cite{crfasrnn_iccv2015,DBLP:journals/corr/LiuLLLT15,DBLP:journals/corr/LinSRH15,Chen2016}, such complex losses often require: 1) modifying the network structure (casting CNN into an RNN)~\cite{crfasrnn_iccv2015,Chen2016}, or 2) using a complicated multi-stage learning scheme, where different layers are optimized during a different training stage~\cite{DBLP:journals/corr/LiuLLLT15,DBLP:journals/corr/LinSRH15}. Due to such complex training procedures, which are adapted for specific tasks and datasets, these models can be quite difficult to adapt for new tasks and datasets, which is disadvantageous.

Inspired by random walk methods~\cite{Lovasz93randomwalks,Bahmani:2010:FIP:1929861.1929864,ilprints422}, in this work we introduce a simple, yet effective alternative to traditional FCNs: a Convolutional Random Walk Network (RWN) that combines the strengths of FCNs and random walk methods. Our model addresses the issues of (1) poor localization around the boundaries suffered by FCNs and (2) spatially disjoint segments produced by dense CRFs. Additionally, unlike recent semantic segmentation approaches~\cite{noh2015learning,crfasrnn_iccv2015,DBLP:journals/corr/LiuLLLT15,DBLP:journals/corr/LinSRH15}, our RWN does so without significantly increasing the complexity of the model. 

Our proposed RWN jointly optimizes (1) pixelwise affinity and (2) semantic segmentation learning objectives that are linked via a novel random walk layer, which enforces spatial consistency in the deepest layers of the network. The random walk layer is implemented via matrix multiplication. As a result, RWN seamlessly integrates both affinity and segmentation branches, and can be jointly trained end-to-end via standard back-propagation with minimal modifications to the existing FCN framework. Additionally, our implementation of RWN requires only $131$ additional parameters. Thus, the effective complexity of our model is the same as the complexity of traditional FCNs (see Table~\ref{complexity_table}). We compare our approach to several variants of the DeepLab semantic segmentation system~\cite{DBLP:journals/corr/ChenPKMY14,DBLP:journals/corr/ChenYWXY15}, and show that our proposed RWN consistently produces better performance over these baselines for the tasks of semantic segmentation and scene labeling.

\captionsetup{labelformat=default}
\captionsetup[figure]{skip=5pt}

 \setlength{\tabcolsep}{3pt}


 \begin{table}[t]
  \footnotesize
    \begin{center}
    \begin{tabular}{| c | c | c | c | c | c | c | c |}
    \hline
          & ~\cite{DBLP:journals/corr/ChenPKMY14} &  ~\cite{Chen2016} & ~\cite{noh2015learning} & ~\cite{crfasrnn_iccv2015} & ~\cite{DBLP:journals/corr/LiuLLLT15}  & \bf RWN\\ \hline
         requires post-processing?  & \cmark & \xmark & \xmark & \xmark & \xmark &\xmark \\ \hline
         uses complex loss? & \xmark & \xmark &  \xmark & \cmark & \cmark &  \xmark\\  \hline
         requires recurrent layers?  &  \xmark & \cmark & \xmark  & \cmark &  \xmark   &\xmark \\ \hline
	model size (in MB) & \bf 79  & \bf 79 & 961  & 514 & >1000  & \bf 79  \\ \hline
    \end{tabular}
    \end{center}\vspace{-.4cm}
    \caption{Summary of recent semantic segmentation models. For each model, we report whether it requires: (1) CRF post-processing, (2) complex loss functions, or (3) recurrent layers. We also list the size of the model (using the size of Caffe~\cite{jia2014caffe} models in MB). We note that unlike prior methods, our RWN does not need post-processing, it is implemented using standard layers and loss functions, and it also has a compact model.\vspace{-0.5cm}}
    \label{complexity_table}
   \end{table}

\begin{figure*}
\begin{center}
   \includegraphics[width=0.8\linewidth]{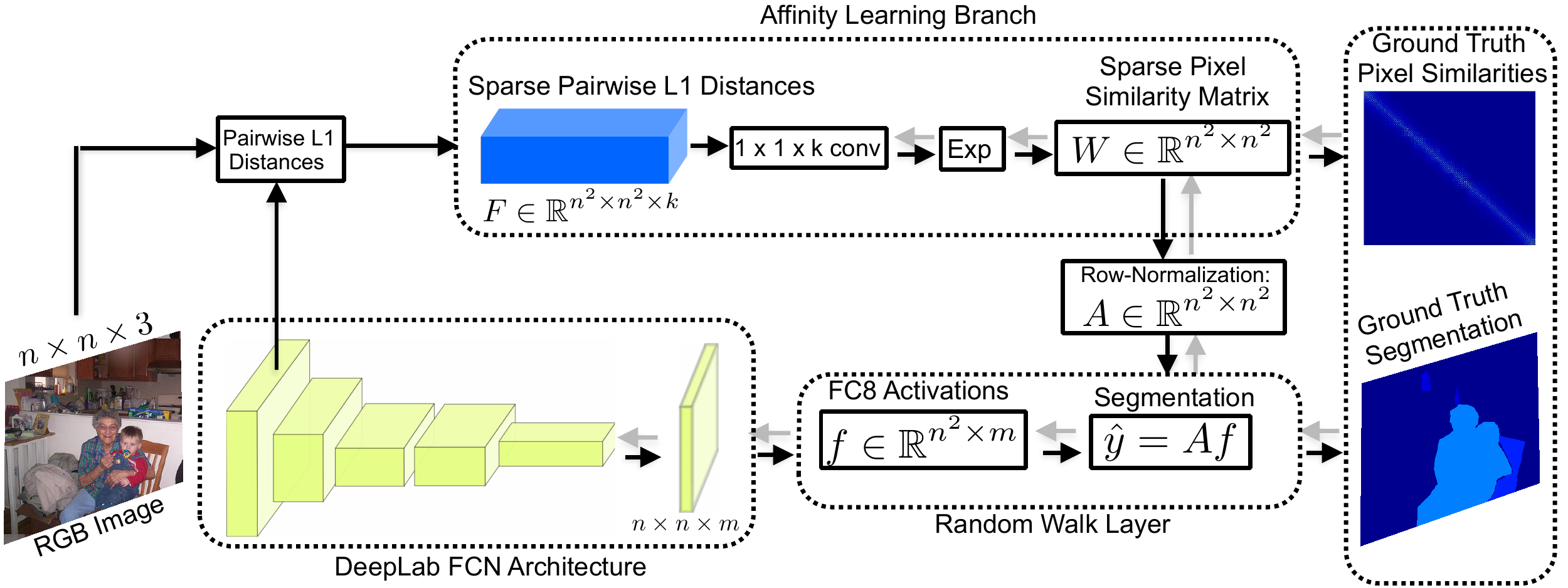}
\end{center}
\setcounter{figure}{1}
   \caption{The architecture of our proposed Random Walk Network (RWN) (best viewed in color). Our RWN consists of two branches: (1) one branch devoted to the segmentation predictions , and (2) another branch predicting pixel-level affinities. These two branches are then merged via a novel random walk layer that encourages  spatially smooth segmentation predictions. The entire RWN is jointly optimized end-to-end via a standard back-propagation algorithm.\vspace{-0.5cm}}
   
\label{fig:arch}
\end{figure*}

\section{Related Work}



The recent introduction of fully convolutional networks (FCNs)~\cite{long_shelhamer_fcn} has led to remarkable advances in semantic segmentation. However, due to the large receptive fields and many pooling layers, segments predicted by FCNs tend to be blobby and lack fine object boundary details. Recently there have been several attempts to address these problems. These approaches can be divided into several groups. 

The work in~\cite{gberta_2016_CVPR,DBLP:journals/corr/ChenPKMY14,Qi_2015_ICCV,DBLP:journals/corr/Kokkinos15,gberta_2016_CVPR} used FCN predictions as unary potentials in a separate globalization model that refines the segments using similarity cues based on regions or boundaries. One disadvantage of these methods is that the learning of the unary potentials and the training of the globalization model are completely disjoint. As a result, these methods often fail to capture semantic relationships between objects, which produces segmentation results that are spatially disjoint (see the right side of Fig.~\ref{intro_fig}).

To address these issues, several recent methods~\cite{crfasrnn_iccv2015,DBLP:journals/corr/LiuLLLT15,DBLP:journals/corr/LinSRH15} have proposed to integrate a CRF or a MRF into the network, thus enabling end-to-end training of the joint model. However, the merging of these two models leads to a dramatic increase in complexity and number of parameters. For instance, the method in~\cite{crfasrnn_iccv2015}, requires to cast the original FCN into a Recurrent Neural Network (RNN), which renders the model much bigger in size (see Table~\ref{complexity_table}). A recent method~\cite{Chen2016} jointly predicts boundaries and segmentations, and then combines them using a recurrent layer, which also requires complex modifications to the existing FCN framework. 

The work in~\cite{DBLP:journals/corr/LiuLLLT15} proposes to use \textbf{local} convolutional layers, which leads to a significantly larger number of parameters. Similarly, the method in~\cite{DBLP:journals/corr/LinSRH15} proposes to model unary and pairwise potentials by separate multi-scale branches. This leads to a network with at least twice as many parameters as the traditional FCN and a much more complex multi-stage training procedure. 


In addition to the above methods, it is worth mentioning deconvolutional networks~\cite{noh2015learning,hong2015decoupled}, which use deconvolution and unpooling layers to recover fine object details from the coarse FCN predictions. However, in order to effectively recover fine details one must employ nearly as many deconvolutional layers as the number of convolutional layers, which yields a large growth in number of parameters (see Table~\ref{complexity_table}). 


Unlike these prior methods, our implementation of RWN needs only $131$ additional parameters over the base FCN. These additional parameters represent only $0.0008 \%$ of the total number of parameters in the network. In addition, our RWN uses standard convolution and matrix multiplication. Thus, it does not need to incorporate complicated loss functions or new complex layers~\cite{crfasrnn_iccv2015,Chen2016,DBLP:journals/corr/LiuLLLT15,DBLP:journals/corr/LinSRH15}. Finally, unlike the methods in~\cite{gberta_2016_CVPR,DBLP:journals/corr/ChenPKMY14,Qi_2015_ICCV,DBLP:journals/corr/Kokkinos15} that predict and refine the segmentations disjointly, our RWN model jointly optimizes pixel affinity and semantic segmentation in an end-to-end fashion. Our experiments show that this leads to spatially smoother segmentations.

\section{Background}

\textbf{Random Graph Walks.} Random walks are one of the most widely known and used methods in graph theory~\cite{Lovasz93randomwalks}. Most notably, the concept of random walks led to the development of PageRank~\cite{ilprints422} and Personalized PageRank~\cite{Bahmani:2010:FIP:1929861.1929864}, which are widely used for many applications. Let $G=(V,E)$ denote an undirected graph with a set of vertices $V$ and a set of edges $E$. Then a random walk in such graph can be characterized by the transition probabilities between its vertices. Let $W$ be a symmetric $n \times n$ affinity matrix, where $n$ denotes the number of nodes in the graph and where $W_{ij} \in [0,1]$ denotes how similar the nodes $i$ and $j$ are. In the context of a semantic segmentation problem, each pixel in the image can be viewed as a separate node in the graph, where the similarity between two nodes can be evaluated according to some metric (e.g. color or texture similarity etc). Then let $D$ indicate a diagonal $n \times n$ matrix, which stores the degree values for each node:  $D_{ii}=\sum_{j=1}^n W_{ij}$ for all $j$ except $i=j$. Then, we can express our random walk transition matrix as $A=D^{-1}W$.

Given this setup, we want to model how the information in the graph spreads if we start at a particular node, and perform a random walk in this graph. Let $y_t$ be a $n \times 1$ vector denoting a node distribution at time $t$. In the context of the PageRank algorithm, $y_t$ may indicate the rank estimates associated with each of the $n$ Web pages at time $t$. Then, according to the random walk theory, we can spread the rank information in the graph by performing a one-step random walk. This process can be expressed as $y_{t+1}=Ay_{t}$, where $y_{t+1}$ denotes a newly obtained rank distribution after one random walk step, the matrix $A$ contains the random walk transition probabilities, and $y_t$ is the rank distribution at time step $t$. Thus, we can observe that the information among the nodes can be diffused, by simply multiplying the random walk transition probability matrix $A$, with the rank distribution $y_t$ at a particular time $t$. This process can be repeated multiple times, until convergence is reached. For a more detailed survey please see~\cite{Lovasz93randomwalks,ilprints422}.

\textbf{Difference from MRF/CRF Approaches.} CRFs and MRFs have been widely used in structured prediction problems~\cite{Lafferty:2001:CRF:645530.655813}. Recently, CRFs and MRFs have also been integrated into the fully convolutional network framework for semantic segmentation~\cite{crfasrnn_iccv2015,DBLP:journals/corr/LiuLLLT15,DBLP:journals/corr/LinSRH15}. We want to stress that while the goals of CRF/MRF and random walk methods are the same (i.e. to globally propagate information in the graph structures), the mechanism to achieve this objective is very different in these two approaches. While MRFs and CRFs typically employ graphs with a fixed grid structure (e.g., one where each node is connected to its four closest neighbors), random walk methods are much more flexible and can implement any arbitrary graph structure via the affinity matrix specification. Thus, since our proposed RWN is based on random walks, it can employ any arbitrary graph structure, which can be beneficial as different problems may require different graph structures.

Additionally, to globally propagate information among the nodes, MRFs and CRFs need to employ approximate inference techniques, because exact inference tends to be intractable in graphs with a grid structure. Integrating such approximate inference techniques into the steps of FCN training and prediction can be challenging and may require lots of domain-specific modifications. In comparison, random walk methods globally propagate information among the nodes via a simple matrix multiplication. Not only is the matrix multiplication efficient and exact, but it is also easy to integrate into the traditional FCN framework for both training and prediction schemes. Additionally, due to the use of standard convolution and matrix multiplication operations, our RWN can be trivially trained via  standard back-propagation in an end-to-end fashion.


\section{Convolutional Random Walk Networks}


In this work, our goal is to integrate a random walk process into the FCN architecture to encourage coherent semantic segmentation among pixels that are similar to each other. Such a process introduces an explicit grouping mechanism, which should be beneficial in addressing the issues of (1) poor localization around the boundaries, and (2) spatially fragmented segmentations.

A schematic illustration of our proposed RWN architecture is presented in Fig.~\ref{fig:arch}. Our RWN is a network composed of two branches: (1) one branch that predicts semantic segmentation potentials, and (2) another branch devoted to predicting pixel-level affinities. These two branches are merged via a novel random walk layer that encourages spatially coherent semantic segmentation. The entire RWN can be jointly optimized end-to-end. We now describe each of the components of the RWN architecture in more detail.



\subsection{Semantic Segmentation Branch}

For the semantic segmentation branch, we present results for several variants of the DeepLab segmentation systems, including DeepLab-LargeFOV~\cite{DBLP:journals/corr/ChenPKMY14}, DeepLab-attention~\cite{DBLP:journals/corr/ChenYWXY15}, and DeepLab-v2, which is one of the top performing segmentation systems. DeepLab-largeFOV is a fully convolutional adaptation of the VGG~\cite{Simonyan14c} architecture, which contains $16$ convolutional layers. DeepLab-attention~\cite{DBLP:journals/corr/ChenYWXY15}, is a multi-scale VGG based network, for which each multi-scale branch focuses on a specific part of the image. Finally, DeepLab-v2 is a multi-scale network based on the residual network~\cite{He_2016_CVPR} implementation. We note that even though we use a DeepLab architecture in our experiments, other architectures such as~\cite{badrinarayanan2015segnet2} and many others could be integrated into our framework.

\subsection{Pixel-Level Affinity Branch}
\label{aff_sec}

To learn the pairwise pixel-level affinities, we employ a separate affinity learning branch with its own learning objective (See Fig.~\ref{fig:arch}). The affinity branch is connected with the input $n \times n \times 3$ RGB image, and low-level $conv1\_1$ and $conv1\_2$ layers. The feature maps corresponding to these layers are $n \times n$  in width and height but they have a different number of channels ($3, 64,$ and $64$ respectively). Let $k$ be the total number of affinity learning parameters (in our case $k=3+64+64=131$). Then, let $F$ be a \textbf{sparse} $n^2 \times n^2 \times k$ matrix that stores $L1$ distances between each pixel and all of its neighbors within a radius $R$, according to each channel. Note that the distances are \textbf{not} summed up across the $k$ channels, but instead they are computed and stored separately for each channel. The resulting matrix $F$ is then used as an input to the affinity branch, as shown in Figure~\ref{fig:arch}. 


The affinity branch consists of a $1 \times 1 \times k$ convolutional layer and an exponential layer. The output of the exponential layer is then attached to the Euclidean loss layer and is optimized to predict the ground truth pixel affinities, which are obtained from the original semantic segmentation annotations. Specifically, we set the ground truth affinity between two pixels to $1$ if the pixels share the same semantic label and have distance less than $R$ from each other.  Note that $F$, which is used as an input to the affinity branch, is a \textbf{sparse} matrix, as only a small fraction of all the entries in $F$ are populated with non-zero values. The rest of the entries are ignored during the computation. 

Also note that we only use features from RGB, $conv1\_1$ and $conv1\_2$ layers, because they are not affected by pooling, and thus, preserve the original spatial resolution.  We also experimented with using features from  deeper FCN layers such as \textit{fc6}, and \textit{fc7}. However, we observed that features from deeper layers are highly correlated to the predicted semantic segmentation unary potentials, which causes redundancy and little improvement in the segmentation performance. We also experimented with using more than one convolutional layer in the affinity learning branch, but observed that additional layers provide negligible improvements in accuracy.

\captionsetup{labelformat=empty}
\captionsetup[figure]{skip=5pt}

\begin{figure}
\centering

%
%



\myfigurethreecol{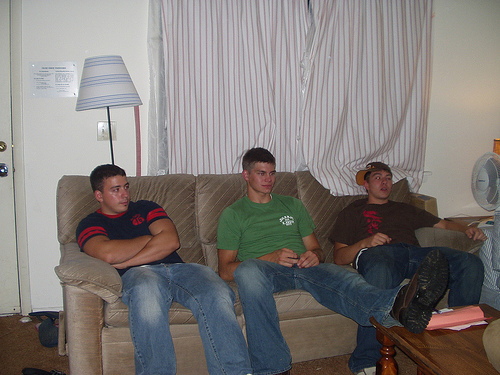}
\myfigurethreecol{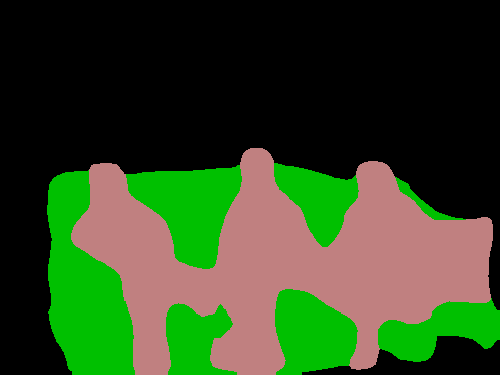}
\myfigurethreecol{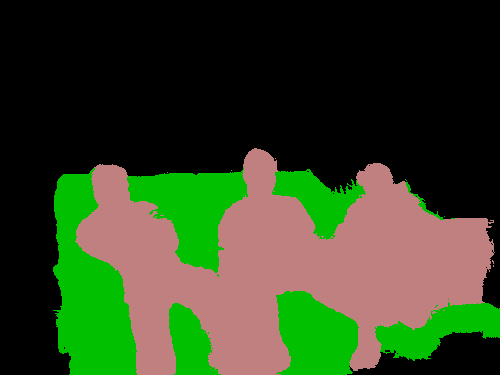}

\myfigurethreecol{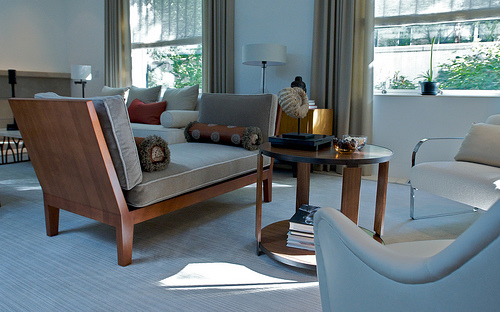}
\myfigurethreecol{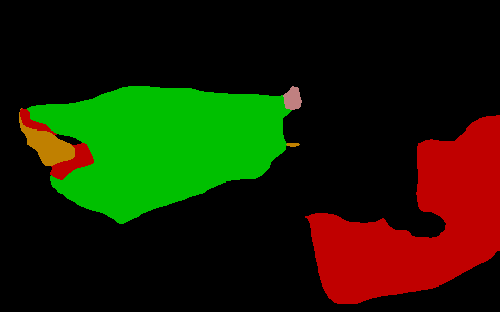}
\myfigurethreecol{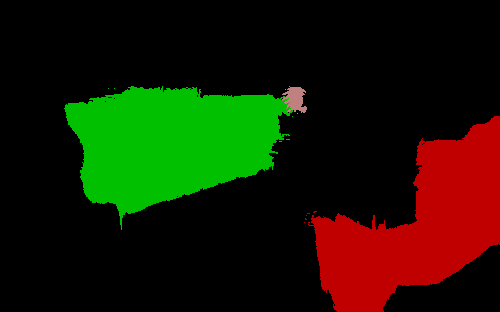}

\myfigurethreecolcaption{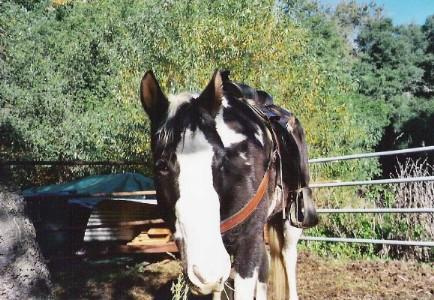}{Input}
\myfigurethreecolcaption{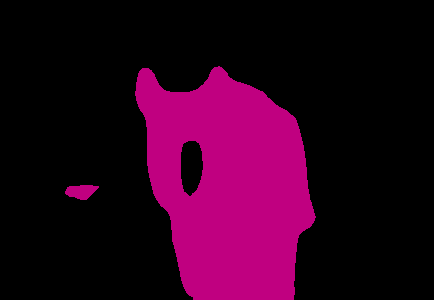}{DeepLab\_v2}
\myfigurethreecolcaption{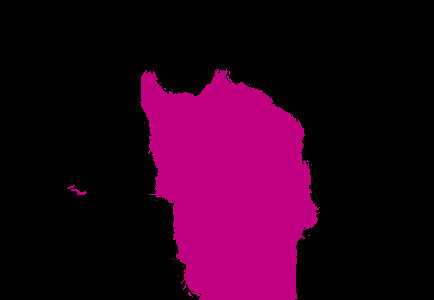}{RWN\_v2}

\captionsetup{labelformat=default}
\setcounter{figure}{2}
    \caption{A figure illustrating the segmentation results of our RWN and the DeepLab-v2 network. Note that RWN produced segmentations are spatially smoother and produce less false positive predictions than the DeepLab-v2 system.\vspace{-0.4cm}}
    \label{dl_fig}
\end{figure}

    \captionsetup{labelformat=default}

  \setlength{\tabcolsep}{2pt}

     \begin{table*}
     \footnotesize
    \begin{center}
    \begin{tabular}{ | c |  c  c  c  c  c  c  c  c  c  c  c  c  c  c  c  c  c  c  c  c ? c  c |}
    \hline
      Method & aero & bike & bird & boat & bottle & bus & car & cat & chair & cow & table & dog & horse & mbike & person & plant & sheep & sofa & train & tv & mean & overall \\ \hline
	 DeepLab-largeFOV & 79.8 & 71.5 & 78.9 & 70.9 & 72.1 & 87.9 & 81.2 & 85.7 & 46.9 & 80.9 & 56.5 & 82.6 & 77.9 & 79.3 & 80.1 & 64.4 & 77.6 & 52.7 & 80.3 & 70.0 &  73.8 & 76.0\\ 
	 \bf RWN-largeFOV & \bf 81.6 & \bf 72.1 & \bf 82.3 & \bf 72.0 & \bf 75.4 & \bf 89.1 & \bf 82.5 & \bf 87.4 & \bf 49.1 & \bf 83.6 & \bf 57.9 & \bf 84.8 & \bf 80.7 & \bf80.2 & \bf 81.2 & \bf 65.7 & \bf 79.7 & \bf 55.5 & \bf  81.5 & \bf 74.0 & \bf 75.8 & \bf 77.9\\ \Xhline{3\arrayrulewidth} 
	 DeepLab-attention & 83.4 & 76.0 & 83.0 & \bf 74.2 & 77.6 & 91.6 & 85.2 & 89.1 & 54.4 & 86.1 & 62.9 & 86.7 & 83.8 & 84.2 & 82.4 & \bf 70.2 & 84.7 & 61.0 & 84.8 & 77.9 &  79.0 & 80.5\\ 
	 \bf RWN-attention & \bf 84.7 & \bf 76.6 & \bf 85.5 & 74.0 & \bf 79.0 & \bf 92.4 & \bf 85.6 & \bf 90.0 & \bf 55.6 & \bf 87.4 & \bf 63.5 & \bf 88.2 & \bf 85.0 & \bf 84.8 & \bf 83.4 & 70.1 & \bf 85.9 & \bf 62.6 & \bf  85.1 & \bf 79.3 & \bf 79.9 & \bf 81.5\\ \Xhline{3\arrayrulewidth} 
	 DeepLab-v2 & 85.5 & \bf 50.6 & 86.9 & \bf 74.4 & 82.7 & 93.1 & 88.4 & 91.9 & 62.1 & 89.7 & 71.5 & 90.3 & 86.2 & 86.3 & 84.6 & \bf 75.1 & 87.6 & 72.2 & 87.8 & 81.3 &  81.4 & 83.4\\ 
	 \bf RWN-v2 & \bf 86.0 & 50.0 & \bf 88.4 & 73.5 & \bf 83.9 & \bf 93.4 & \bf 88.6 & \bf 92.5 & \bf 63.9 & \bf 90.9 & \bf 72.6 & \bf 90.9 & \bf 87.3 & \bf 86.9 & \bf 85.7 & 75.0 & \bf 89.0 & \bf 74.0 & \bf  88.1 & \bf 82.3 & \bf 82.1 & \bf 84.3\\ \hline 
      \end{tabular}
    \end{center}
    \vspace{-0.4cm}
        \caption{Semantic segmentation results on the SBD dataset according to the per-pixel Intersection over Union evaluation metric. From the results, we observe that our proposed RWN consistently outperforms DeepLab-LargeFOV, DeepLab-attention, and DeepLab-v2 baselines. \vspace{-0.5cm}}
    \label{seg_table}
   \end{table*}

\subsection{Random Walk Layer}
\label{rw_sec}

To integrate the semantic segmentation potentials and our learned pixel-level affinities, we introduce a novel random walk layer, which propagates the semantic segmentation information based on the learned affinities. The random walk layer is connected to the two bottom layers: (1) the \textit{fc8} layer containing the semantic segmentation potentials, and (2) the affinity layer that outputs a sparse $n^2 \times n^2$ random walk transition matrix $A$. Then, let $f$ denote the activation values from the \textit{fc8} layer, reshaped to the dimensions of $n^2 \times m$, where $n^2$ refers to the number of pixels, and $m$ is the number of object classes in the dataset. A single random walk layer implements one step of the random walk process, which can be performed as $\hat{y}=Af$, where $\hat{y}$ indicates the diffused segmentation predictions, and $A$ denotes the random walk transition matrix. 

The random walk layer is then attached to the softmax loss layer, and is optimized to predict ground truth semantic segmentations. One of the advantages of our proposed random walk layer is that it is implemented as a matrix multiplication, which makes it possible to back-propagate the gradients to both (1) the affinity branch and (2) the segmentation branch. Let the softmax-loss gradient be an $n^2 \times m$ matrix $\frac{\partial{L}}{\partial{\hat{y}}}$, where $n^2$ is the number of pixels in the \textit{fc8} layer, and $m$ is the number of predicted object classes. Then the gradients, which are back-propagated to the semantic segmentation branch are computed as $\frac{\partial{L}}{\partial{f}}=A^T\frac{\partial{L}}{\partial{\hat{y}}}$, where $A^T$ is the transposed random walk transition matrix. Also, the gradients, that are back-propagated to the affinity branch are computed as $\frac{\partial{L}}{\partial{A}}=\frac{\partial{L}}{\partial{\hat{y}}}f^T$, where $f^{T}$ is a $m \times n^2$ matrix that contains transposed activation values from the \textit{fc8} layer of the segmentation branch. We note that $\frac{\partial{L}}{\partial{A}}$ is a \textbf{sparse} $n^2 \times n^2$ matrix, which means that the above matrix multiplication only considers the pixel pairs that correspond to the non-zero entries in the random walk transition matrix $A$. 


\subsection{Random Walk Prediction at Testing}

In the previous subsection, we mentioned that the prediction in the random walk layer can be done via a simple matrix multiplication operation $\hat{y}=Af$, where $A$ denotes the random walk transition matrix, and $f$ depicts the activation values from the \textit{fc8} layer. Typically, we want to apply multiple random walk steps until convergence is reached. However, we also do not want to deviate too much from our initial segmentation predictions, in case the random walk transition matrix is not perfectly accurate, which is a reasonable expectation. Thus, our prediction scheme needs to balance two effects: (1) propagating the segmentation information across the nodes using a random walk transition matrix, and (2) not deviating too much from the initial segmentation. 

This tradeoff between the two quantities is very similar to the idea behind MRF and CRF models, which try to minimize an energy formed by unary and pairwise terms. However, as discussed earlier, MRF and CRF methods  tend to use 1) grid structure graphs and 2) various approximate inference techniques to propagate segmentation information globally. In comparison, our random walk approach is advantageous because it can use 1) any arbitrary graph structure and 2) an exact matrix multiplication operation to achieve the same goal.

Let us first denote our segmentation prediction after $t+1$ random walk steps as $\hat{y}^{t+1}$. Then our general prediction scheme can be written as:

\begin{equation} \label{eq:pred1}
\hat{y}^{t+1}=\alpha A\hat{y}^{t}+(1-\alpha)f
\end{equation}

where $\alpha$ denotes a parameter $\interval{0}{1}$ that controls the tradeoff between (1) diffusing segmentation information along the connections of a random walk transition matrix and (2) not deviating too much from initial segmentation values (i.e. the outputs of the last layer of the FCN). Let us now initialize  $\hat{y}^0$ to contain the output values from the \textit{fc8} layer, which we denoted with $f$. Then we can write our prediction equation by substituting the recurrent expressions:

%
%

\begin{equation} \label{eq:pred2}
\hat{y}^{t+1}=(\alpha A)^{t+1}f+(1-\alpha)\sum_{i=0}^{t}(\alpha A)^if
\end{equation}

Now, because we want to apply our random walk procedure until convergence we set $t=\infty$. Then, because our random walk transition matrix is stochastic we know that $\lim_{t \to \infty} (\alpha A)^{t+1} = 0$. Furthermore, we can  write $S_t=\sum_{i=0}^{t}A^i=I+A+A^2+\hdots+A^t$, where $I$ is an identity matrix, and where $S_t$ denotes a partial sum of random walk transitions until iteration $t$. We can then write $S_t-AS_t=I-A^{t+1}$, which implies: 

\begin{equation} \label{eq:pred3}
(I-A)S_t=I-A^{t+1}
\end{equation}

From our previous derivation, we already know that  $\lim_{t \to \infty} (A)^{t+1} = 0$, which implies that

\begin{equation} \label{eq:pred4}
S_\infty=(I-A)^{-1}
\end{equation}

Thus, our final prediction equation, which corresponds to applying repeated random walk steps until convergence, can be written as 

\begin{equation} \label{eq:pred5}
\hat{y}^{\infty}=(I-\alpha A)^{-1}f
\end{equation}

In practice, the random walk transition matrix $A$ is pretty large, and inverting it is impractical. To deal with this problem, we shrink the matrix $(I-\alpha A)$ using a simple and efficient technique presented in~\cite{cArbelaez14}, and then invert it to compute the final segmentation. In the experimental section, we show that such a prediction scheme produces solid results and is still pretty efficient ($\approx$ 1 second per image). We also note that we use this prediction scheme only during testing. During training we employ a scheme that uses a single random walk step (but with a much larger radius), which is faster. We explain this procedure in the next subsection.

\captionsetup{labelformat=empty}
\captionsetup[figure]{skip=5pt}

\begin{figure}
\centering

\myfigurethreecol{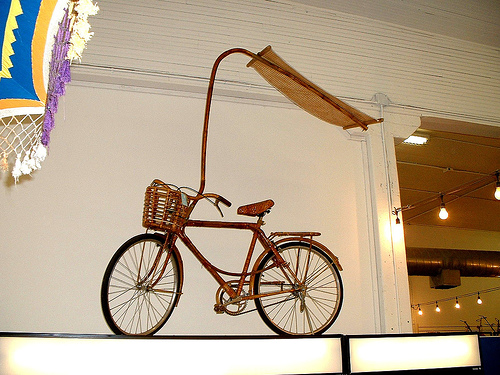}
\myfigurethreecol{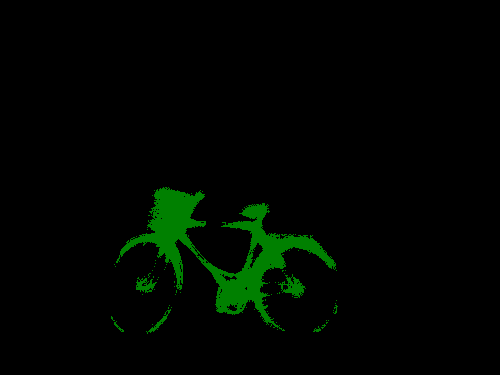}
\myfigurethreecol{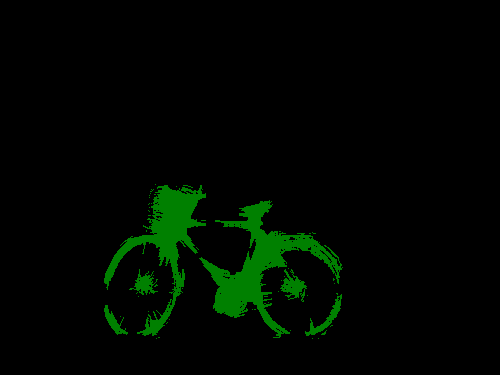}

\myfigurethreecolcaption{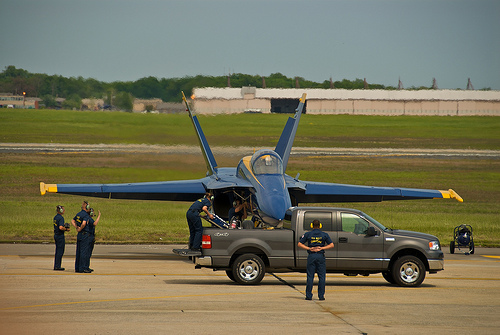}{Input}
\myfigurethreecolcaption{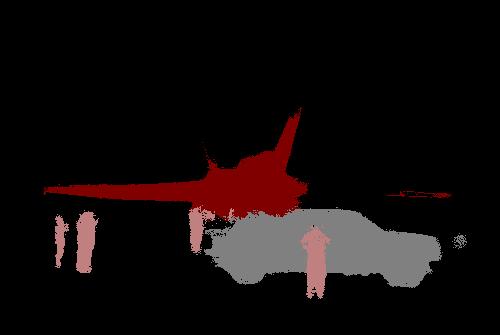}{DeepLab\_v2-CRF}
\myfigurethreecolcaption{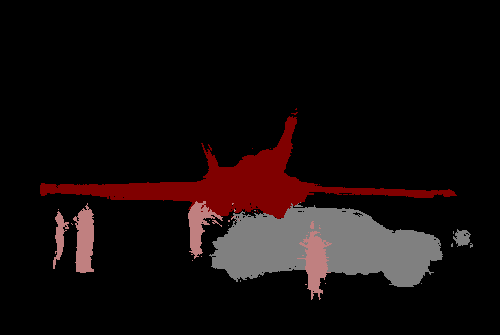}{RWN\_v2}


\captionsetup{labelformat=default}
\setcounter{figure}{3}
    \caption{Comparison of segmentation results produced by our RWN versus the DeepLab-v2-CRF system. It can be noticed that, despite not using any post-processing steps, our RWN predicts fine object details (e.g., bike wheels or plane wings) more accurately than DeepLab-v2-CRF, which fails to capture some these object parts.\vspace{-0.4cm}} 
    \label{crf_fig}
\end{figure}

\subsection{Implementation Details}
\label{impl_sec}


We jointly train our RWN in an end-to-end fashion for $2000$ iterations, with a learning rate of $10^{-5}$, $0.9$ momentum, the weight decay of $5 \cdot 10^{-5}$, and $15$ samples per batch. For the RWN model, we set the tradeoff parameter $\alpha$ to $0.01$. During testing we set the random walk connectivity radius $R=5$ and apply the random walk procedure until convergence. However, during training we set $R=40$, and apply a single random walk step. This training strategy works well because increasing the radius size eliminates the need for multiple random walk steps, which speeds up the training. However, using $R=5$ and applying an infinite number of random walk steps until convergence still yields slightly better results (see study in~\ref{abl_exp}), so we use it during testing. For all of our experiments, we use a Caffe library~\cite{jia2014caffe}. During training, we also employ data augmentation techniques such as cropping, and mirroring.  





\section{Experimental Results}



In this section, we present our results for semantic segmentation on the SBD~\cite{BharathICCV2011} dataset, which contains objects and their per-pixel labels for $20$ Pascal VOC classes (excluding the background class). We also include scene labeling results on the commonly used Stanford background~\cite{Gould+al:ICCV09} and Sift Flow~\cite{Liu2016} datasets. We evaluate our segmentation results on these tasks using the standard metric of the intersection-over-union (IOU) averaged per pixels across all the classes from each dataset. We also include the class-agnostic overall pixel intersection-over-union score, which measures the per-pixel IOU across all classes.


We experiment with several variants of the DeepLab system~\cite{DBLP:journals/corr/ChenPKMY14,DBLP:journals/corr/ChenYWXY15} as our main baselines throughout our experiments: DeepLab-LargeFOV~\cite{DBLP:journals/corr/ChenPKMY14}, DeepLab-attention~\cite{DBLP:journals/corr/ChenYWXY15}, and DeepLab-v2. 

   \setlength{\tabcolsep}{2.5pt}
\captionsetup{labelformat=default}
\begin{table}
   \small
    \begin{center}
    \begin{tabular}{ | c  | c | c |}
    \hline
     Method & mean IOU & overall IOU\\ \hline
	DeepLab-largeFOV-CRF &  75.7 & 77.7\\
	RWN-largeFOV & \bf 75.8 & \bf 77.9 \\ \hline 
	DeepLab-attention-CRF &  \bf 79.9 & \bf 81.6\\ 
	RWN-attention & \bf 79.9 & 81.5\\ \hline 
	DeepLab-v2-CRF &  81.9 & 84.2\\ 
	RWN-v2 & \bf 82.1 & \bf 84.3\\ \hline 
	DeepLab-DT & 76.6 & 78.7\\ 
	RWN & \bf 76.7 & \bf 78.8\\ \hline
    \end{tabular}
    \end{center}\vspace{-.2cm}
    \caption{Quantitative comparison between our RWN model and several variants of the DeepLab system that use a dense CRF or a domain-transfer (DT) filter for post-processing. These results suggest that our RWN acts as an effective globalization scheme, since it produces results that are similar or even better than the results achieved by post-processing the DeepLab outputs with a CRF or DT.\vspace{-0.4cm}}
    \label{crf_table}
   \end{table}


Our evaluations provide evidence for four conclusions:


\begin{itemize}
\item In subsections~\ref{seg_exp},~\ref{scene_exp}, we demonstrate  that our proposed RWN outperforms DeepLab baselines for both semantic segmentation, and scene labeling tasks.
\item In subsection~\ref{seg_exp}, we demonstrate that, compared to the dense CRF approaches, RWN predicts segmentations that are spatially smoother.
\item In Subsection~\ref{runtime_sec}, we show that our approach is more efficient than the denseCRF inference.
\item Finally, in Subsection,~\ref{abl_exp}, we demonstrate that our random walk layer is beneficial and that our model is flexible to use different graph structures.
\end{itemize}


\subsection{Semantic Segmentation Task}
\label{seg_exp}

\textbf{Standard Evaluation.} In Table~\ref{seg_table}, we present semantic segmentation results on the Pascal SBD dataset~\cite{BharathICCV2011}, which contains $8055$ training and $2857$ testing images. These results indicate that RWN consistently outperforms all three of the DeepLab baselines. In Figure~\ref{dl_fig}, we also compare qualitative segmentation results of a DeepLab-v2 network and our RWN model. We note that the RWN segmentations contain fewer false positive predictions and are also spatially smoother across the object regions.


\captionsetup{labelformat=default}

 \begin{figure}
\begin{center}
   \includegraphics[width=0.65\linewidth]{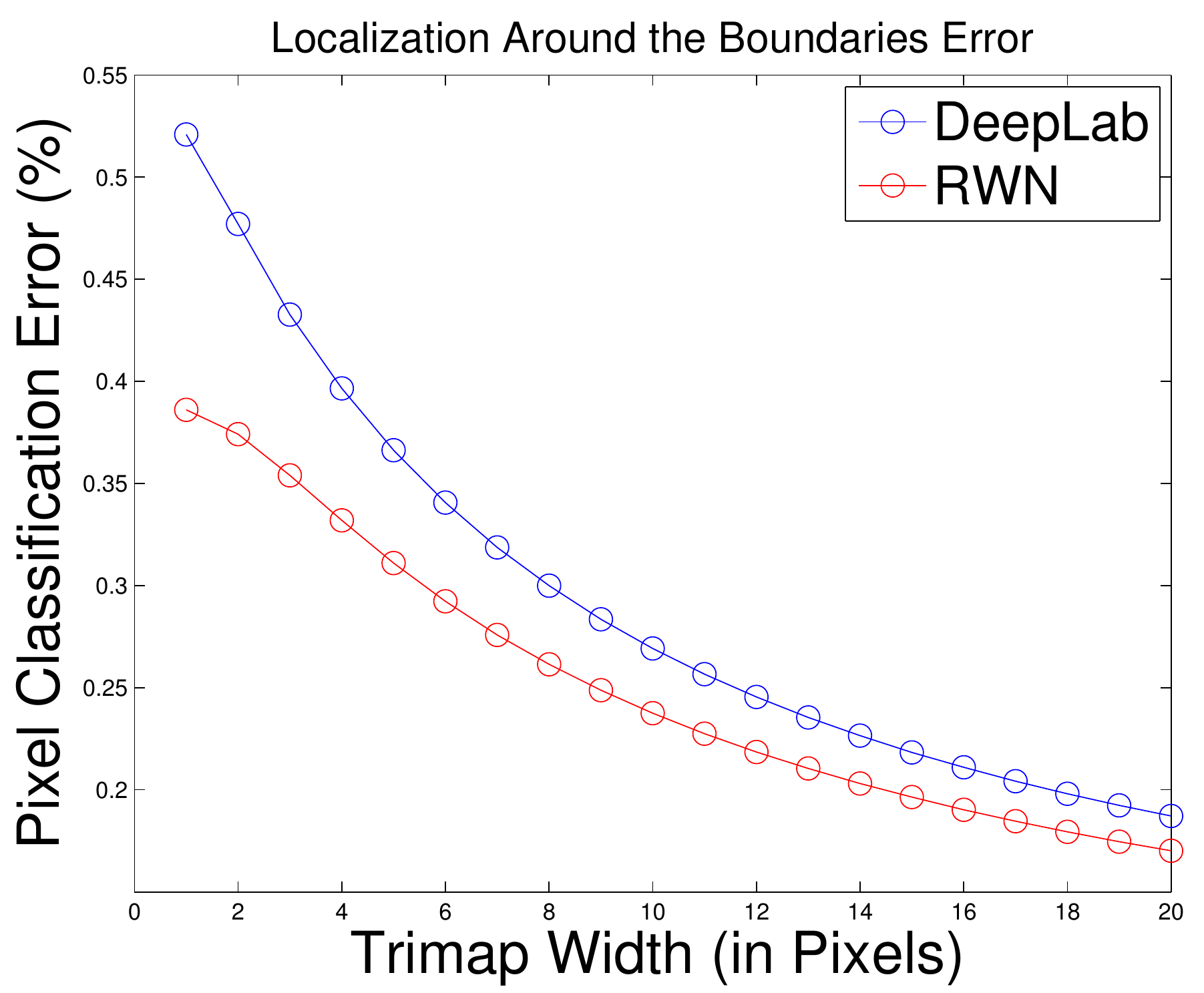}
\end{center}
   \caption{Localization error around the object boundaries within a trimap. Compared to the DeepLab system (blue), our RWN (red) achieves lower segmentation error around object boundaries for all trimap widths. \vspace{-0.3cm}}
   
\label{trimap}
\end{figure}


Furthermore, in Table~\ref{crf_table}, we present experiments where we compare RWN with models using dense CRFs~\cite{NIPS2011_4296} to post-process the predictions of DeepLab systems. We also include DeepLab-DT~\cite{Chen2016}, which uses domain-transfer filtering to refine the segmentations inside an FCN. Based on these results, we observe that, despite not using any post-processing, our RWN produces results similar to or even better than the DeepLab models employing post-processing. These results indicate that RWN can be used as a globalization mechanism to ensure spatial coherence in semantic segmentation predictions. In Figure~\ref{crf_fig} we  present qualitative results where we compare the final segmentation predictions of RWN and the DeepLab-v2-CRF system. Based on these qualitative results, we observe that RWN captures more accurately the fine details of the objects, such as the bike wheels, or plane wings.  The DeepLab-v2-CRF system misses some of these object parts.



\textbf{Localization Around the Boundaries.} Earlier we claimed that due to the use of large receptive fields and many pooling layers, FCNs tend to produce blobby segmentations that lack fine object boundary details. We want to show that our RWN produces more accurate segmentations around object boundaries the traditional FCNs. Thus,  adopting the practice from~\cite{NIPS2011_4296}, we evaluate segmentation accuracy around object boundaries. We do so by counting the relative number of misclassified pixels within a narrow
band (``trimap'') surrounding the ground truth object boundaries. We present these results in Figure~\ref{trimap}. The results show that RWN achieves higher segmentation accuracy than the DeepLab (DL) system for all trimap widths considered in this test.

\textbf{Spatial Smoothness.} We also argued that applying the dense CRF~\cite{NIPS2011_4296} as a post-processing technique often leads to spatially fragmented segmentations (see the right side of Fig.~\ref{intro_fig}). How can we evaluate whether a given method produces spatially smooth or spatially fragmented segmentations? Intuitively, spatially fragmented segmentations will produce many false boundaries that do not correspond to  actual object boundaries. Thus, to test the spatial smoothness of a given segmentation, we extract the boundaries from the segmentation and then compare these boundaries against ground truth object boundaries using the standard maximum F-score (MF) and average precision (AP) metrics, as done in the popular BSDS benchmark~\cite{MartinFTM01}. We perform this experiment on the  Pascal SBD dataset and present these results in Table~\ref{bsds_table}. We can see that the boundaries extracted from the RWN segmentations yield better MF and AP results compared to the boundaries extracted from the different variants of the DeepLab-CRF system. Thus, these results suggest that RWN produces spatially smoother segmentations than DeepLab-CRF.

\setlength{\tabcolsep}{2.5pt}
\captionsetup{labelformat=default}
\begin{table}
   \small
    \begin{center}
    \begin{tabular}{ | c  | c | c |}
    \hline
     Method & MF & AP\\ \hline
	DeepLab-largeFOV-CRF  & 0.676 & 0.457 \\
	RWN-largeFOV & \bf 0.703 & \bf 0.494\\ \hline 
	DeepLab-attention-CRF & 0.722 & 0.521\\ 
	RWN-attention & \bf 0.747 & \bf 0.556\\ \hline 
	DeepLab-v2-CRF  & 0.763 & 0.584\\ 
	RWN-v2 & \bf 0.773 & \bf 0.595 \\ \hline 

    \end{tabular}
    \end{center}\vspace{-.2cm}
    \caption{Quantitative comparison of spatial segmentation smoothness. We extract the boundaries from the predicted segmentations and evaluate them against ground truth object boundaries using max F-score (MF) and average precision (AP) metrics. These results suggest that RWN segmentations are spatially smoother than the DeepLab-CRF segmentations across all baselines.\vspace{-0.4cm}}
    \label{bsds_table}
   \end{table}

\subsection{Scene Labeling}
\label{scene_exp}

We also tested our RWN on the task of scene labeling using two popular datasets: Stanford Background~\cite{Gould+al:ICCV09} and Sift Flow~\cite{Liu2016}. Stanford Background is a relatively small dataset for scene labeling. It contains $715$ images, which we randomly split into $600$ training images and $115$ testing images. In contrast, the Sift Flow dataset contains $2489$ training examples and $201$ testing images. For all of our experiments, we use the DeepLab-largeFOV~\cite{DBLP:journals/corr/ChenPKMY14} architecture since it is smaller and more efficient to train and test. To evaluate scene labeling results, we use the overall IOU evaluation metric which is a commonly used metric for this task. In Table~\ref{scene_table}, we present our scene labeling results on both of these datasets. Our results indicate that our RWN method outperforms the DeepLab baseline by $2.57 \%$, and $2.54\%$ on these two datasets, respectively.
 

\begin{figure}
\begin{center}
   \includegraphics[width=1\linewidth]{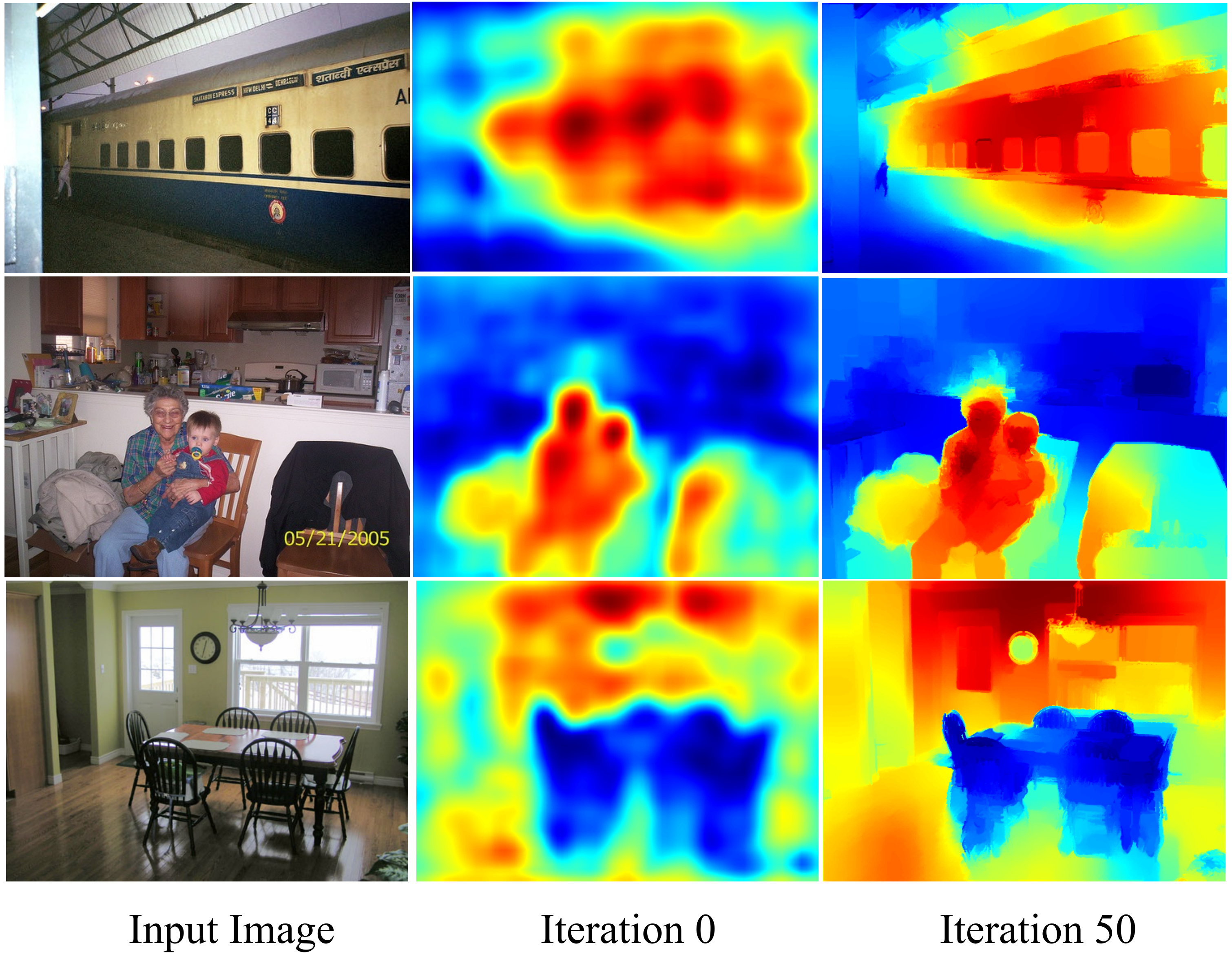}
\end{center}
  \vspace{-0.3cm}
   \caption{A figure illustrating how the probability predictions change as we apply more random walk steps. Note that the RWN predictions become more refined and better localized around the object boundaries as more random walk steps are applied.\vspace{-0.5cm}}
   
\label{fig:evolution}
\end{figure}

\subsection{Runtime Comparisons}
\label{runtime_sec}

We also include the runtime comparison of our RWN approach versus the denseCRF inference. We note that using a single core of a $2.7$GHz Intel Core i7 processor, the denseCRF inference requires 3.301 seconds per image on average on a Pascal SBD dataset. In comparison, a single iteration of a random walk, which is simply a sparse matrix multiplication, takes 0.032 seconds on average on the same Pascal SBD dataset.  A DeepLab\_v2 post-processed with denseCRF achieves $81.9 \%$ IOU score on this same Pascal SBD dataset. In comparison, RWN\_v2 with a single random walk iteration and with R=40 (radius) achieves $82.2 \%$ IOU, which is both more accurate and more than $100$ times more efficient than the denseCRF inference.

\subsection{Ablation Experiments}
\label{abl_exp}

\begin{figure}
\begin{center}
   \includegraphics[width=0.65\linewidth]{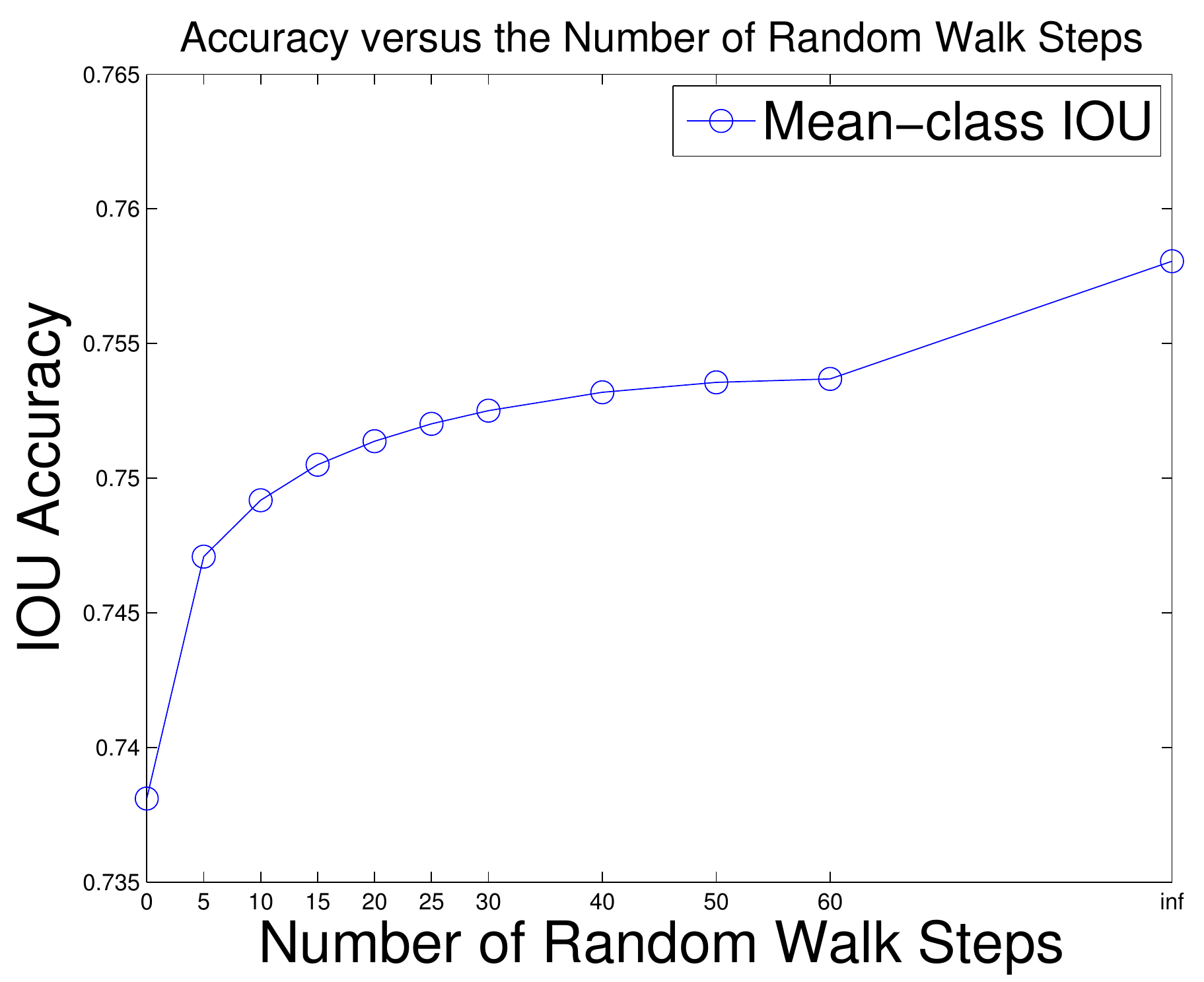}
\end{center}
   \caption{IOU accuracy as a function of the number of random walk steps. From this plot we observe that the segmentation accuracy keeps improving as we apply more random walk steps and that it reaches its peak when the random walk process converges.\vspace{-0.3cm}}
   
\label{fig:rw_steps}
\end{figure}

\textbf{Optimal Number of Random Walk Steps.} In Figure~\ref{fig:rw_steps}, we illustrate how the IOU accuracy changes when we use a different number of random walk steps. We observe that the segmentation accuracy keeps increasing as we apply more random walk steps, and that it reaches its peak performance when the random walk process converges, which indicates the effectiveness of our random walk step procedure. In Figure~\ref{fig:evolution}, we also illustrate how the predicted object segmentation probabilities change as we apply more random walk steps. We observe that the object boundaries become much better localized as more iterations of random walk are applied.



\textbf{Radius Size.} To analyze the effect of a radius size in the RWN architecture, we test alternative versions of our model with different radii sizes. Our results indicate, that the RWN model produces similar results with different radii in the interval of $R>3$ and $R<20$ if the random walk step process is applied until convergence. We also note that, if we select $R=40$, and apply a random walk step only \textbf{once}, we can achieve the segmentation accuracy of $75.5\%$ and $77.6\%$ according to the two evaluation metrics, respectively. In comparison, choosing $R=5$ and applying random walk until convergence yields the accuracies of $75.8\%$ and $77.9\%$, which is slightly better. However, note that selecting $R=40$, and applying multiple random walk steps does not yield any improvement in segmentation accuracy. These experiments show the flexibility of our model compared to the MRF or CRF models, which typically use graphs with a fixed grid structure. Our model has the ability to use different graph structures depending on the problem.

\section{Conclusion}

In this work, we introduced Random Walk Networks (RWNs), and showed that, compared to traditional fully convolutional networks (FCNs), they produce improved accuracy for the same model complexity. Our RWN addresses the issues of  1) poor localization around the segmentation boundaries and 2) spatially disjoint segmentations. Additionally, our implementation of RWN uses only $131$ additional learnable parameters ($0.0008\%$ of the original number of the parameters in the network) and it can be easily integrated into the standard FCN learning framework for a joint end-to-end training. Finally, RWN provides a more efficient alternative to the denseCRF approaches.



Our future work includes experimenting with alternative RWN architectures and applying RWN to new domains such as language processing or speech recognition.

\captionsetup{labelformat=default}

   \captionsetup{labelformat=default}
  \begin{table}
   \small
    \begin{center}
    \begin{tabular}{  c | c |  c |}
    \cline{2-3}
    & \multicolumn{1}{ c |}{DeepLab-largeFOV} & \multicolumn{1}{ c |}{RWN-largeFOV} \\ \cline{1-3}
      \multicolumn{1}{| c |}{Stanford-BG} & 65.74  & \bf 68.31 \\ 
      \multicolumn{1}{| c |}{Sift-Flow} & 67.31  & \bf 69.85 \\ \hline
    \end{tabular}
    \end{center}\vspace{-.4cm}
    \caption{Scene labeling results on the Stanford Background and Sift-Flow datasets measured according to the overall IOU evaluation metric.  We use a DeepLab-largeFOV network as base model, and show that our RWN yields better results on both of these scene labeling datasets.\vspace{-0.3cm}}
    \label{scene_table}
   \end{table}

\bibliographystyle{plain}
\small{
\bibliography{gb_bibliography}}

\begin{thebibliography}{10}

\bibitem{cArbelaez14}
Pablo Arbelaez, J.~Pont-Tuset, Jon Barron, F.~Marqu{\'e}s, and Jitendra Malik.
\newblock Multiscale combinatorial grouping.
\newblock In {\em Computer Vision and Pattern Recognition (CVPR)}, 2014.

\bibitem{badrinarayanan2015segnet2}
Vijay Badrinarayanan, Ankur Handa, and Roberto Cipolla.
\newblock Segnet: A deep convolutional encoder-decoder architecture for robust
  semantic pixel-wise labelling.
\newblock {\em arXiv preprint arXiv:1505.07293}, 2015.

\bibitem{Bahmani:2010:FIP:1929861.1929864}
Bahman Bahmani, Abdur Chowdhury, and Ashish Goel.
\newblock Fast incremental and personalized pagerank.
\newblock {\em Proc. VLDB Endow.}, 4(3):173--184, December 2010.

\bibitem{gberta_2016_CVPR}
Gedas Bertasius, Jianbo Shi, and Lorenzo Torresani.
\newblock Semantic segmentation with boundary neural fields.
\newblock In {\em The IEEE Conference on Computer Vision and Pattern
  Recognition (CVPR)}, June 2016.

\bibitem{Chen2016}
Liang{-}Chieh Chen, Jonathan~T. Barron, George Papandreou, Kevin Murphy, and
  Alan~L. Yuille.
\newblock Semantic image segmentation with task-specific edge detection using
  cnns and a discriminatively trained domain transform.
\newblock {\em CVPR}, 2016.

\bibitem{DBLP:journals/corr/ChenPKMY14}
Liang{-}Chieh Chen, George Papandreou, Iasonas Kokkinos, Kevin Murphy, and
  Alan~L. Yuille.
\newblock Semantic image segmentation with deep convolutional nets and fully.
\newblock In {\em ICLR}, 2015.

\bibitem{DBLP:journals/corr/ChenYWXY15}
Liang{-}Chieh Chen, Yi~Yang, Jiang Wang, Wei Xu, and Alan~L. Yuille.
\newblock Attention to scale: Scale-aware semantic image segmentation.
\newblock {\em CVPR}, 2016.

\bibitem{DBLP:journals/corr/DaiH015}
Jifeng Dai, Kaiming He, and Jian Sun.
\newblock Boxsup: Exploiting bounding boxes to supervise convolutional networks
  for semantic segmentation.
\newblock In {\em The IEEE International Conference on Computer Vision (ICCV)},
  December 2015.

\bibitem{Gould+al:ICCV09}
S.~Gould, R.~Fulton, and D.~Koller.
\newblock Decomposing a scene into geometric and semantically consistent
  regions.
\newblock In {\em Proceedings of the International Conference on Computer
  Vision (ICCV)}, 2009.

\bibitem{BharathICCV2011}
Bharath Hariharan, Pablo Arbelaez, Lubomir Bourdev, Subhransu Maji, and
  Jitendra Malik.
\newblock Semantic contours from inverse detectors.
\newblock In {\em International Conference on Computer Vision (ICCV)}, 2011.

\bibitem{He_2016_CVPR}
Kaiming He, Xiangyu Zhang, Shaoqing Ren, and Jian Sun.
\newblock Deep residual learning for image recognition.
\newblock In {\em The IEEE Conference on Computer Vision and Pattern
  Recognition (CVPR)}, June 2016.

\bibitem{hong2015decoupled}
Seunghoon Hong, Hyeonwoo Noh, and Bohyung Han.
\newblock Decoupled deep neural network for semi-supervised semantic
  segmentation.
\newblock In {\em NIPS)}, December 2015.

\bibitem{jia2014caffe}
Yangqing Jia, Evan Shelhamer, Jeff Donahue, Sergey Karayev, Jonathan Long, Ross
  Girshick, Sergio Guadarrama, and Trevor Darrell.
\newblock Caffe: Convolutional architecture for fast feature embedding.
\newblock {\em arXiv preprint arXiv:1408.5093}, 2014.

\bibitem{DBLP:journals/corr/Kokkinos15}
Iasonas Kokkinos.
\newblock Surpassing humans in boundary detection using deep learning.
\newblock {\em CoRR}, abs/1511.07386, 2015.

\bibitem{NIPS2011_4296}
Philipp Kr\"{a}henb\"{u}hl and Vladlen Koltun.
\newblock Efficient inference in fully connected crfs with gaussian edge
  potentials.
\newblock In J.~Shawe-Taylor, R.S. Zemel, P.L. Bartlett, F.~Pereira, and K.Q.
  Weinberger, editors, {\em Advances in Neural Information Processing Systems
  24}, pages 109--117. Curran Associates, Inc., 2011.

\bibitem{Lafferty:2001:CRF:645530.655813}
John~D. Lafferty, Andrew McCallum, and Fernando C.~N. Pereira.
\newblock Conditional random fields: Probabilistic models for segmenting and
  labeling sequence data.
\newblock In {\em Proceedings of the Eighteenth International Conference on
  Machine Learning}, ICML '01, pages 282--289, San Francisco, CA, USA, 2001.
  Morgan Kaufmann Publishers Inc.

\bibitem{DBLP:journals/corr/LinSRH15}
Guosheng Lin, Chunhua Shen, Ian~D. Reid, and Anton van~den Hengel.
\newblock Efficient piecewise training of deep structured models for semantic
  segmentation.
\newblock {\em CoRR}, abs/1504.01013, 2015.

\bibitem{Liu2016}
Ce~Liu, Jenny Yuen, and Antonio Torralba.
\newblock {\em Nonparametric Scene Parsing via Label Transfer}, pages 207--236.
\newblock Springer International Publishing, Cham, 2016.

\bibitem{DBLP:journals/corr/LiuLLLT15}
Ziwei Liu, Xiaoxiao Li, Ping Luo, Chen~Change Loy, and Xiaoou Tang.
\newblock Semantic image segmentation via deep parsing network.
\newblock In {\em ICCV}, 2015.

\bibitem{long_shelhamer_fcn}
Jonathan Long, Evan Shelhamer, and Trevor Darrell.
\newblock Fully convolutional networks for semantic segmentation.
\newblock In {\em The IEEE Conference on Computer Vision and Pattern
  Recognition (CVPR)}, June 2015.

\bibitem{Lovasz93randomwalks}
Laszlo Lovasz.
\newblock Random walks on graphs: A survey, 1993.

\bibitem{MartinFTM01}
D.~Martin, C.~Fowlkes, D.~Tal, and J.~Malik.
\newblock A database of human segmented natural images and its application to
  evaluating segmentation algorithms and measuring ecological statistics.
\newblock In {\em Proc. 8th Int'l Conf. Computer Vision}, volume~2, pages
  416--423, July 2001.

\bibitem{noh2015learning}
Hyeonwoo Noh, Seunghoon Hong, and Bohyung Han.
\newblock Learning deconvolution network for semantic segmentation.
\newblock In {\em Computer Vision (ICCV), 2015 IEEE International Conference
  on}, 2015.

\bibitem{ilprints422}
Lawrence Page, Sergey Brin, Rajeev Motwani, and Terry Winograd.
\newblock The pagerank citation ranking: Bringing order to the web.
\newblock Technical Report 1999-66, Stanford InfoLab, November 1999.
\newblock Previous number = SIDL-WP-1999-0120.

\bibitem{Qi_2015_ICCV}
Xiaojuan Qi, Jianping Shi, Shu Liu, Renjie Liao, and Jiaya Jia.
\newblock Semantic segmentation with object clique potential.
\newblock In {\em The IEEE International Conference on Computer Vision (ICCV)},
  December 2015.

\bibitem{Simonyan14c}
K.~Simonyan and A.~Zisserman.
\newblock Very deep convolutional networks for large-scale image recognition.
\newblock {\em CoRR}, abs/1409.1556, 2014.

\bibitem{crfasrnn_iccv2015}
Shuai Zheng, Sadeep Jayasumana, Bernardino Romera-Paredes, Vibhav Vineet,
  Zhizhong Su, Dalong Du, Chang Huang, and Philip Torr.
\newblock Conditional random fields as recurrent neural networks.
\newblock In {\em International Conference on Computer Vision (ICCV)}, 2015.

\end{thebibliography}

\end{document}